\documentclass[runningheads]{llncs}
\usepackage{graphicx}
\usepackage{cite}
\usepackage{latexsym}
\usepackage{amsmath,amssymb,amsfonts}
\usepackage[linesnumbered,ruled]{algorithm2e}
\usepackage{booktabs}
\usepackage{multirow}
\usepackage{floatrow}
\newfloatcommand{capbtabbox}{table}[][\FBwidth]
\usepackage{amssymb}
\usepackage{float}
\floatstyle{plaintop}
\restylefloat{table}

\usepackage[pagebackref,breaklinks,colorlinks]{hyperref}
\hypersetup{
        final,
        breaklinks,
        colorlinks,
        linkcolor={red!69!black},
        citecolor={green!69!black},
        urlcolor={blue!69!black}
        }

\usepackage{tikz}
\usetikzlibrary{shapes}
\newcommand{\markerone}{\raisebox{0.5pt}{\tikz{\node[draw,scale=0.4,circle,fill=white](){};}}}
\newcommand{\markertwo}{\raisebox{0.5pt}{\tikz{\node[draw,scale=0.4,circle,fill=black!20!white](){};}}}
\newcommand{\markerthree}{\raisebox{0.5pt}{\tikz{\node[draw,scale=0.4,regular polygon, regular polygon sides=4,fill=none](){};}}}
\newcommand{\markerfour}{\raisebox{0.5pt}{\tikz{\node[draw,scale=0.3,regular polygon, regular polygon sides=3,fill=none,rotate=0](){};}}}

\begin{document}
	\title{Anatomy-Aware Contrastive Representation Learning for Fetal Ultrasound}
	%
	%
\author{
Zeyu Fu\inst{1}\thanks{Equal contribution.} \and
Jianbo Jiao\inst{1}\protect\footnotemark[1] \and
Robail Yasrab\inst{1}\protect\footnotemark[1]\and
Lior Drukker\inst{2,3}\and
Aris T. Papageorghiou\inst{2}\and
J. Alison Noble\inst{1}}
\authorrunning{Z. Fu et al.}
\institute{
Department of Engineering Science, University of Oxford, Oxford, UK \\ \email{zeyu.fu@eng.ox.ac.uk}
\and
Nuffield Department of Women's and Reproductive Health, University of Oxford, Oxford, UK
\and 
Department of Obstetrics and Gynecology, Tel-Aviv University, Israel}
\maketitle              
\begin{abstract}		
Self-supervised contrastive representation learning offers the advantage of learning meaningful visual representations from unlabeled medical datasets for transfer learning. 
However, applying current contrastive learning approaches to medical data without considering its domain-specific anatomical characteristics may lead to visual representations that are inconsistent in appearance and semantics. 
In this paper, we propose to improve visual representations of medical images via anatomy-aware contrastive learning (AWCL), which incorporates anatomy information to augment the positive/negative pair sampling in a contrastive learning manner. 
The proposed approach is demonstrated for automated fetal ultrasound imaging tasks, enabling the positive pairs from the same or different ultrasound scans that are anatomically similar to be pulled together and thus improving the representation learning. 
We empirically investigate the effect of inclusion of anatomy information with coarse- and fine-grained granularity, for contrastive learning and find that learning with fine-grained anatomy information which preserves intra-class difference is more effective than its counterpart. We also analyze the impact of anatomy ratio on our AWCL framework and find that using more distinct but anatomically similar samples to compose positive pairs results in better quality representations.
Experiments on a large-scale fetal ultrasound dataset demonstrate that our approach is effective for learning representations that transfer well to three clinical downstream tasks,
and achieves superior performance compared to ImageNet supervised and the current state-of-the-art contrastive learning methods. 
In particular, AWCL outperforms ImageNet supervised method by 13.8\% and state-of-the-art contrastive-based method by 7.1\% on a cross-domain segmentation task. 
\end{abstract}
	
\section{Introduction}
\label{introduction}
Semi-supervised and self-supervised representation learning with or without annotations have attracted significant attention across various medical imaging modalities \cite{model-genesis,Semantic-genesis,C2L,USCL,Rubic-cube}. 
These learning schemes are able to well exploit large-scale unlabeled medical datasets and learn  meaningful representations for downstream task finetuning. 
In particular,  contrastive representation learning based on instance discrimination tasks \cite{simclr, MoCo} has become the leading paradigm for self-supervised pretraining, 
where a model is trained to pull together each instance and its augmented views and meanwhile push it away from those of all other instances in the embedding space. 
\begin{figure*}[t]
		\centering
		\includegraphics[width=1\linewidth]{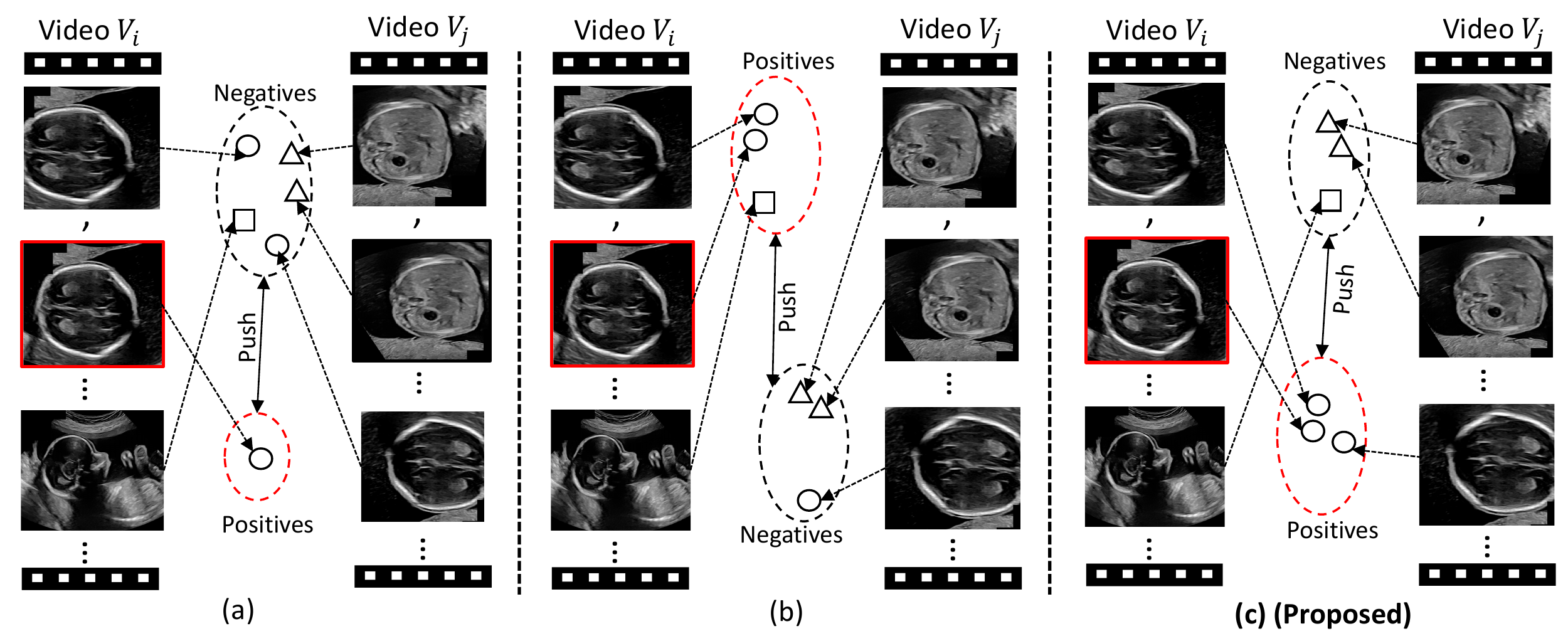}
		\caption{
			Illustration of different representation learning approaches for fetal ultrasound, (a) self-supervised contrastive learning, (b) contrastive learning with patient metadata, and (c) our proposed anatomy-aware contrastive learning. 
			Icon shapes of circle (\protect\markerone), square (\protect\markerthree) and triangle (\protect\markerfour) denote the anatomical categories of fetal head, profile, and abdomen, respectively.  The anchor image is highlighted with a red bounding box, while the red dotted circle means pull together (Best viewed in colored version).}
		\label{fig:justification}
\end{figure*}

However, directly applying self-supervised contrastive learning (e.g. SimCLR \cite{simclr} and MoCo \cite{MoCo}) in the context of medical imaging may result in visual representations that are inconsistent in appearance and semantics.
We illustrate this issue in Fig. \ref{fig:justification}(a), which shows that a vanilla contrastive learning approach without considering the domain-specific anatomical characteristics leads to false negatives, i.e. some negative samples having high affinity with the anchor image are ``pushed away".
To address this, we explore the following question: \emph{Is domain-specific anatomy information helpful in learning better representations for medical data?}

We investigate this question via the proposed anatomy-aware contrastive learning (AWCL), as depicted in Fig. \ref{fig:justification}(c), where ``anatomy-aware" here means that the inclusion of anatomy information is leveraged to augment the positive/negative pair sampling in a contrastive learning manner.
In this work, we demonstrate the proposed approach for fetal ultrasound imaging tasks, where a number of different fetal anatomies can be present in a diagnostic scan.  
Motivated by Khosla et al. \cite{SCL}, we expand the pool of positive samples by grouping images from the same or different ultrasound scans that share common anatomical categories.
More importantly, our approach is optimized alternately with both conventional and anatomy-aware contrastive objectives, as shown in Fig. \ref{fig:overview}(a), given that the anatomy information is not always accessible for each sampling process.
Moreover, we consider both coarse- and fine-grained anatomical categories with the availability for data sampling, as shown in Fig.~\ref{fig:overview}(b) and (c). We also empirically investigate their effect on the transferability of the learned feature representations.
To assess the effectiveness of our pre-trained representations, we evaluated transfer learning on three downstream clinical tasks: standard plane detection, segmentation of Crown Rump Length (CRL) and
Nuchal Translucency (NT), and recognition of first-trimester anatomical structures.
In summary, the main contributions and findings are:
\begin{itemize}
    \item We develop an anatomy-aware contrastive learning approach for medical fetal ultrasound imaging tasks. 
    \item We empirically compare the effect of inclusion of anatomy information with coarse- and fine-grained granularity respectively, within our contrastive learning approach. 
The comparative analysis suggests that contrastive learning with fine-grained anatomy information which preserves intra-class difference is more effective than its counterpart.
    \item  Experimental evaluations on three downstream clinical tasks demonstrate the better generalizability of our proposed approaches over learning from an ImageNet pre-trained ResNet, vanilla contrastive learning \cite{simclr}, and contrastive learning with patient information \cite{USCL,MedAug,Big-self-supervised}.
    \item  We provide an in-depth analysis to show the proposed approach learns high-quality discriminative representations.
\end{itemize}

\section{Related work} 
\subsubsection{Self-supervised learning (SSL) in medical imaging.}
Prior works using SSL for medical imaging typically selecting on designing pre-text tasks, such as solving a Rubik' cube \cite{Rubic-cube}, image restoration \cite{model-genesis,DICOM}, predicting anatomical position \cite{position-prediction} and multi-task joint reasoning~\cite{jiao2020self}.
Recently, contrastive based SSL \cite{simclr, MoCo} has been favourably applied to learn more discriminative representations across various medical imaging tasks \cite{C2L, USCL, MOCO-CXR}.
In particular, Sowrirajan et al.  \cite{MOCO-CXR} successfully adapted a MoCo-contrastive learning method \cite{MoCo} into chest X-rays and demonstrated better transferable representations and initialization for chest X-ray diagnostic tasks.
Taher et al. \cite{TL-Medical-benchmark} presented a benchmark evaluation study to investigate the effectiveness of several established contrastive learning models pre-trained on ImageNet on a variety of medical imaging tasks. 
In addition, there have been recent approaches  \cite{USCL,MedAug,Big-self-supervised}  that leverage patient metadata to improve the medical imaging contrastive learning. 
These approaches constrain the selection of positive pairs only from the same subject (video), with the assumption that visual representations from the same subject share similar semantic meaning. 
However, these approaches may not generalize well to a scenario, where different organs or anatomical structures are captured in a single video. 
For instance, as seen from Fig. \ref{fig:justification}(b), some positive pairs having low affinity in visual appearance and semantics are pulled together, i.e. false positives, which can degrade the representation learning. 
The proposed learning scheme, as shown in Fig. \ref{fig:justification}(c), is advantageous to address the aforementioned limitations by augmenting the sampling process with the inclusion of anatomy information. Moreover, our approach differs from \cite{C2L} and \cite{USCL} which combine label information as an additional supervision signal with self supervision for multi-tasking.

\subsubsection{Representation learning in fetal ultrasound.} 
There are related works exploring representation learning for fetal ultrasound imaging tasks. 
Baumgartner et al. \cite{sononet}  and Schlemper et al. \cite{attention-gated} proposed a VGG-based network and an attention-gated network respectively to detect fetal standard planes. 
Sharma et al. \cite{sharma2021knowledge} presented a multi-stream network which combines 2D image and spatio-temporal information to automate clinical workflow description of full-length routine fetal anomaly ultrasound scans. 
Cai et al. \cite{cai2020multi-task}  considered incorporating the temporal dimension into visual attention modelling via multi-task learning for standard biometry plane-finding navigation.
However, the generalization and transferability of those models to other target tasks remains unclear. 
Droste et al. \cite{Richard_ipmi} proposed to learn transferable representations for fetal ultrasound interpretation  by modelling sonographer visual attention (gaze tracking) without manual supervision. 
More recently, Jiao et al. \cite{Jianbo2020miccai}  proposed to derive a meaningful representation from raw data by developing a cross-modal contrastive learning which aligns the correspondence between fetal ultrasound video and narrative speech audio.
Our work differs by focusing on learning general image representations without requiring additional data modalities (e.g. gaze tracking and audio) from the domain of interest, and we also perform extensive experimental analysis on three downstream clinical tasks to assess the effectiveness of the learned representations. 

\section{Fetal Ultrasound Imaging Dataset}
\label{Fetal Ultrasound Imaging Dataset}
This study uses a large-scale fetal ultrasound imaging dataset , which was acquired as part of PULSE (Perception Ultrasound by Learning Sonographic Experience) project \cite{PULSE-DATASET}. 
The scans were performed by operators including sonographers and fetal medicine specialists using a commercial Voluson E8 version BT18 (General Electric, Zipf, Austria) ultrasound machine. 
During a routine scanning session, the operator views several fetal or maternal anatomical structures. 
The frozen views saved by sonographers are referred to as \textit{standard planes} in the paper, following the UK Fetal Anomaly Screening Programme (FASP) nomenclature \cite{FASP}. 
	
Fetal ultrasound videos, recorded from the ultrasound scanner display using a lossless compression and sampled at the rate of 30 Hz. We consider a subset of the entire ultrasound dataset for the proposed pre-training approach. This consists of total number of 2,050,432  frames\footnote{\scriptsize Every 8th frame is extracted to reduce temporal redundancy of ultrasound videos.} from 534 second-trimester ultrasound videos. 
In this sub-dataset, there are 15,384 frames labeled with 13 fine-grained anatomy categories, including four views of heart, three-vessel and trachea (3VT), four-chamber (4CH), right ventricular outflow tract (RVOT), and left ventricular outflow tract (LVOT), two views of brain, transventricular (BrainTv.) and transcerebellum (BrainTc.), two views of spine, coronal (SpineCor.) and sagittal (SpineSag.), abdomen, femur, kidneys, lips, profile and background class. 
In addition, 69,671 frames are labeled with coarse anatomy categories without dividing the heart, brain and spine into further sub-categories as those of above, but also 3D mode, maternal anatomy including Doppler, abdomen, nose and lips,  kidneys, face-side profile, full-body-side profile, bladder including Doppler, femur and ``Other" class. All image frames were preprocessed by cropping the ultrasound image region and resizing it to 224$\times$288 pixels.

\begin{figure}[t]
	\centering
	\includegraphics[width=1\linewidth]{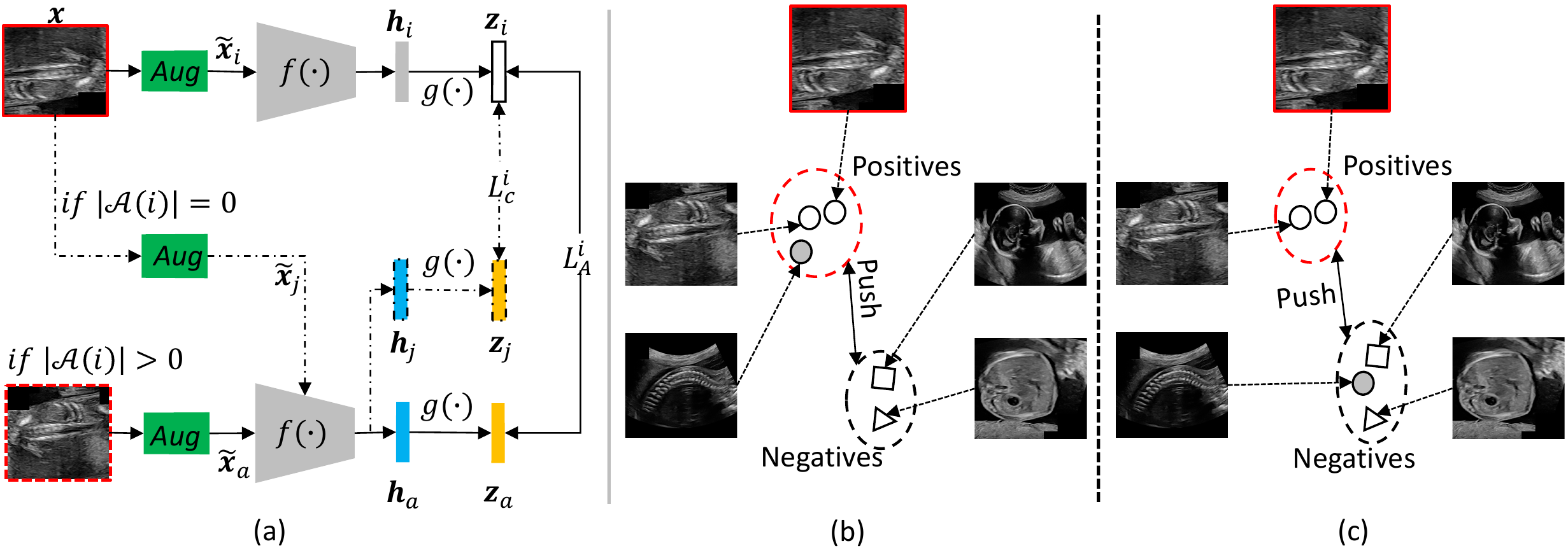}
		\caption{
			(a) presents the overview of proposed anatomy-aware contrastive learning approach. (b) and (c) illustrate using coarse and fine-grained anatomy categories, respectively for the proposed AWCL framework. Icon shapes of white-circle (\protect\markerone), grey-circle (\protect\markertwo), square (\protect\markerthree) and triangle (\protect\markerfour) denote the classes of coronal view of spine, sagittal view of spine, profile, and abdomen, respectively. }
		\label{fig:overview}
\end{figure}

\section{Method}
\label{Method}
In this section, we first describe the problem formulation of contrastive learning with medical images, and then present our anatomy-aware contrastive learning algorithm design as well as training details.

\subsection{Problem formulation} 
\label{Problem Formulation}
For each input image $\mathbf{x}$ in a mini-batch of $N$ samples, randomly sampled from a pre-training dataset $\mathcal{V}$, a contrastive learning framework (i.e. SimCLR \cite{simclr}) applies two augmentations to obtain a positive pair $(\tilde{\mathbf{x}}_{i}, \tilde{\mathbf{x}}_{j})$, yielding a set of $2N$ samples. Let $i$ denote the anchor input index, the contrastive learning objective can be defined as,
\begin{equation}\label{eq:simclr}
	L_{C}^{i}=-\log \frac{\exp \left(\operatorname{sim}\left(\mathbf{z}_{i}, \mathbf{z}_{j}\right) / \tau\right)}{\sum_{k=1}^{2 N} \mathbf{1}_{[k \neq i]} \exp \left(\operatorname{sim}\left(\mathbf{z}_{i}, \mathbf{z}_{k}\right) / \tau\right)},
\end{equation}
where $\mathbf{1}\in\{0,1\}$, $\tau$ is a temperature parameter and $sim(\cdot)$ is the pairwise cosine similarity.
$\mathbf{z}$ is a representation vector, calculated by $\mathbf{z}= g(f(\mathbf{x}))$, where $f(\cdot)$ denotes a shared encoder modelled by a convolutional neural network (CNN) and $g(\cdot)$ is a multi-layer perception (MLP) projection head.

The above underpins the vanilla contrastive learning. However in some cases (e.g. ultrasound scan as illustrated in this paper), this standard approach, as well as its extended version that leverages patient information \cite{USCL,MedAug,Big-self-supervised},  may lead to false negatives and false positives respectively, as seen from Fig. \ref{fig:justification}(a) and (b). To this end, we introduce a new approach as detailed next.

\subsection{Anatomy-aware contrastive learning} 
\label{AWCL}
Fig. \ref{fig:justification}(c) illustrates the main idea of the new anatomy-aware contrastive learning (AWCL) approach, which incorporates additional samples belonging to the same anatomy category from the same or different US scans. In addition to positive sampling from the same image and its augmentation, AWCL is tailored to the case where multiple anatomical structures are present. 

As shown in Fig. \ref{fig:overview}(a), we utilize the available anatomy information as detailed in Section \ref{Fetal Ultrasound Imaging Dataset}, forming a positive sample set $\mathcal{A}(i)$ with the same anatomy as sample $i$. 
The assumption for such a design is that image samples within the same anatomy category should have similar appearances, based on a clinical perspective \cite{PULSE-DATASET}.
Motivated by~\cite{SCL}, we design the anatomy-aware contrastive learning objective as follows,
\begin{equation}\label{eq:aac}
		L_{A}^{i}=-\frac{1}{|\mathcal{A}(i)|}\sum_{a\in\mathcal{A}(i)}\log \frac{\exp \left(\operatorname{sim}\left(z_{i}, z_{a}\right) / \tau\right)}{\sum_{k=1}^{2 N} \mathbf{1}_{[k \neq i]} \exp \left(\operatorname{sim}\left(z_{i}, z_{k}\right) / \tau\right)},
	\end{equation}
where $|\mathcal{A}(i)| $ denotes the cardinality.

Due to the limited availability of some anatomical categories, $\mathcal{A}(i)$ is not always achievable for each sampling process. 
In this regard, the AWCL framework is formulated as an alternate optimization combining both learning objectives of
Eq. \ref{eq:simclr}  and Eq. \ref{eq:aac}. This gives a loss function defined as
\begin{equation}\label{eq:loss}
L^{i}= 
	\begin{cases}
		L_{C}^{i}& \text { if } |\mathcal{A}(i)| = 0 \\ 
L_{A}^{i}& \text { if } |\mathcal{A}(i)| > 0.
	\end{cases}
\end{equation}

Furthermore, we consider both coarse- and fine-grained anatomical categories for the proposed AWCL framework, and compare their effect on the transferability of visual representations. 
Fig. \ref{fig:overview}(b) and (c) shows the motivation of this comparative analysis. 
For an anatomical structure with different views of visual appearance (e.g. the spine has two views as sub-classes), we observe that AWCL with coarse-grained anatomy information tends to minimize the intra-class difference by pulling together all the instances of the same anatomy. 
In contrast, AWCL with fine-grained anatomy information tends to preserve the intra-class difference by pushing away images with different visual appearances despite the same anatomy. 
Both strategies of the proposed learning approach are evaluated and compared in Section \ref{Analysis of Data granularity in AWCL}.
We further study the impact of the ratio of anatomy information used in AWCL pre-training in Section  \ref{Impact of label fraction in AWCL}.

 \begin{algorithm}[t]
	\SetKwInOut{Input}{Input}
	\SetKwInOut{Output}{Output}
	
	\Input{Sample $x_i$ and its positive set $\mathcal{A}(i)$, pre-training dataset $\mathcal{V}$}
	\Output{The loss value $L$ of the current learning step}
	{Sample mini-batch training data $x_i\in\mathcal{V}$}\\
	\eIf{$|\mathcal{A}(i)| = 0$}
	{
	    Apply data augmentations $\rightarrow$ positive pair $(\tilde{\mathbf{x}}_{i}, \tilde{\mathbf{x}}_{j})$\\
	    Extract representation vectors $z_i=g(f(\tilde{\mathbf{x}}_{i})), z_j=g(f(\tilde{\mathbf{x}}_{j}))$\\
		$L=-\log \frac{\exp \left(\operatorname{sim}\left(\mathbf{z}_{i}, \mathbf{z}_{j}\right) / \tau\right)}{\sum_{k=1}^{2 N} \mathbf{1}_{[k \neq i]} \exp \left(\operatorname{sim}\left(\mathbf{z}_{i}, \mathbf{z}_{k}\right) / \tau\right)}$\\
	}
	{
	    Sample data $x_a$ with the same anatomy as $x_i$, where $x_a\in \mathcal{A}(i)$\\
	    Apply data augmentations $\rightarrow$ positive pair $(\tilde{\mathbf{x}}_{i}, \tilde{\mathbf{x}}_{a})$\\
	    Extract representation vectors $z_i=g(f(\tilde{\mathbf{x}}_{i})), z_a=g(f(\tilde{\mathbf{x}}_{a}))$\\
		$L=-\frac{1}{|\mathcal{A}(i)|}\sum_{a\in\mathcal{A}(i)}\log \frac{\exp \left(\operatorname{sim}\left(z_{i}, z_{a}\right) / \tau\right)}{\sum_{k=1}^{2 N} \mathbf{1}_{[k \neq i]} \exp \left(\operatorname{sim}\left(z_{i}, z_{k}\right) / \tau\right)}$\\
	}
	Return $L$
	\caption{Anatomy-aware Contrastive Learning (AWCL)}\label{algorithm-awcl}
\end{algorithm}
\subsection{Implementation details} 
\label{AWCL-settings}
Algorithm \ref{algorithm-awcl} provides the pseudo-code of AWCL.
Following the prior art \cite{USCL,MedAug,MOCO-CXR}, we use ResNet-18 \cite{resnet} as our backbone architecture. Further studies on different network architectures are out of scope of this paper.  
We split the pre-training dataset as detailed in Section \ref{Fetal Ultrasound Imaging Dataset} into training and validation sets (80\%/20\%), and train the model using the Adam optimizer with a weight decay of $10^{-6}$, and a mini-batch size of 32. 
We follow~\cite{simclr} for the data augmentations applied to the sampled training data. The output feature dimension of $z$ is set to 128. The temperature parameter $\tau$ is set as 0.5.
The models are trained with the loss functions defined earlier (Eq.~\ref{eq:aac} and Eq.~\ref{eq:simclr}) for 10 epochs. The learning rate is set as $10^{-3}$. 
The whole framework is implemented with the PyTorch \cite{NEURIPS2019_9015} framework on a PC with NVIDIA Titan V GPU card. The code is available at \url{https://github.com/JianboJiao/AWCL}.

To demonstrate the effectiveness of AWCL trained models, we compare them with random initialization, ImageNet pre-trained ResNet18 \cite{resnet},  supervised pre-training with coarse labels, supervised pre-training with fine-grained labels, vanilla contrastive learning (SimCLR) \cite{simclr},  and contrastive learning with patient information (CLPI) \cite{USCL,Big-self-supervised,CLOCS}. 
All pre-training methods presented here are pre-trained from scratch on the pre-training dataset with the similar parameter configurations as listed above.

\section{Experiments on Transfer Learning}
\label{Transfer Learning}
In this section, we evaluate the effectiveness of the SSL pre by supervised transfer learning with end-to-end fine-tuning on three downstream clinical tasks, which are second-trimester standard plane detection (Task I), recognition of first-trimester anatomies (Task II) and segmentation of NT and CRL (Task III). 
The datasets for downstream task evaluation are listed in Table \ref{tab:data}, and are independent datasets from \cite{PULSE-DATASET}. 
For fair comparison, all compared pre-training models were fine-tuned with the same parameter settings and data augmentation policies within each downstream task evaluation. 

\subsection{Evaluation on standard plane detection}
\label{Evaluation on Standard Plane Detection}
\subsubsection{Evaluation details.} 
Here, we investigate how the pre-trained representations generalize to an in-domain second-trimester classification task, which consists of the same fine-grained anatomical categories as detailed in Section \ref{Fetal Ultrasound Imaging Dataset}. 
We fine-tune each trained backbone encoder and attach a classifier head \cite{sononet} to train the entire network for 70 epochs with a learning rate of 0.01, decayed by 0.1 at epochs 30 and 55. 
The network training is performed via SGD with momentum of 0.9, weight decay of 5×$10^{-4}$, mini-batch size of 16 and a cross-entropy loss, and it is evaluated via a three-fold cross validation. 
The augmentation policy used is analogous to \cite{Richard_ipmi}, including random horizontal flipping, rotation (10 degrees), and varying gamma and brightness. 
We employ precision, recall and F1-scores computed as macro-averages as the evaluation metrics. 
\begin{table}[t]
	\caption{\footnotesize Details of the downstream datasets and imaging tasks.}
	\centering
	\label{tab:data}
	\resizebox{0.8\linewidth}{!}{%
		\begin{tabular}{l|l|c|c|c}
			\toprule
			Trimester&Task&\#Scans & \#Images&  \#Classes   \\
			\hline
			2nd& I- Standard Plane Detection& 58 &   1,075& 14 \\
			1st& II- Recognition of first-trimester anatomies&90  & 25,490 &5\\
			1st& III- Segmentation of NT and CRL & 128 & 16,093 &3 \\
			\bottomrule
		\end{tabular}
	}	
\end{table}
\subsubsection{Results and discussion.} 
\begin{table}[t]
	\caption{\footnotesize Quantitative comparison of fine-tuning performance (mean±std.[\%]) on the tasks of standard plane detection (Task I), first-trimester anatomy recognition (Task II) and CRL / NT segmentation (Task III). Best results are
		marked in \textbf{bold.}}
	\centering
	\label{tab:result}
	\resizebox{1\linewidth}{!}{
		\begin{tabular}{l|ccc|ccc|ccc}
			\toprule
			&\multicolumn{3}{@{}c@{}}{Task I}&   \multicolumn{3}{@{}c@{}}{Task II}&   \multicolumn{3}{@{}c@{}}{Task III}\\
			\hline
			Pre-training methods  &Precision ($\uparrow$) & Recall ($\uparrow$) &   F1-score ($\uparrow$) &   Precision ($\uparrow$)&   Recall ($\uparrow$)&   F1-score ($\uparrow$)&
			GAA ($\uparrow$)&MA($\uparrow$)&mIoU($\uparrow$)
			\\
			\hline
			Rand.Init.&70.4$\pm$1.7&   58.3$\pm$3.1 & 61.6$\pm$3.1&
			81.4$\pm3.4$ &79.2$\pm0.1$ &81.5$\pm0.3$& 
			67.3$\pm$0.2 &63.0$\pm2.1$ &46.7$\pm$0.1                    \\
			
			ImageNet &78.8$\pm$4.6 &73.6$\pm$4.1  &73.6$\pm$2.8&
			92.0$\pm0.5$ &93.4$\pm$1.5 &92.1$\pm$2.9 & 
			71.6$\pm1.3$ &64.2 $\pm$ 1.0 & 49.0$\pm$0.1			\\
			
			Supervised (coarse)& 74.2$\pm$2.7&  67.5$\pm$3.4 &69.0$\pm$3.2&
			95.2$\pm0.1$&93.7$\pm0.2$ &94.1$\pm0.4$& 
			76.4$\pm0.3$ &67.5$\pm1.1$ &50.1$\pm0.3$ 			\\
			
			Supervised (fine-grained) & \textbf{84.6$\pm$1.0}&  \textbf{77.1$\pm$2.3}&\textbf{78.6$\pm$2.1}& 
			96.1$\pm0.1$&96.8$\pm1.0$ &96.4$\pm0.9$& 
			80.0$\pm0.2$ &75.5$\pm0.1$ &62.8$\pm0.4$ 			\\
			SimCLR &71.7$\pm$0.3 & 69.6$\pm$1.5  &69.4$\pm$0.7& 
			96.0$\pm0.5$& 95.2$\pm0.3$ &95.2$\pm0.8$&
			77.6$\pm$1.4& 69.2$\pm$0.1 &55.7$\pm0.2$ \\
			
			CLPI &68.6$\pm$4.2 &68.5$\pm$3.2 &67.5$\pm$3.7 &
			89.2$\pm$0.1&88.3$\pm$0.8 &89.6$\pm$1.1 &
			72.7$\pm$0.2&65.4$\pm$1.4& 48.1$\pm$1.2\\		
	
			\hline
			\textit{AWCL (coarse)} &71.4$\pm$3.3& 73.1$\pm$1.9 &71.3$\pm$2.2& 95.6$\pm$0.7&96.2$\pm$1.6&95.9$\pm$0.2 &
			79.8$\pm$0.7&76.1$\pm$0.3&61.2$\pm$1.3\\
			\textit{AWCL (fine-grained)} &71.8$\pm$2.7 &70.0$\pm$1.2   & 70.1$\pm$1.7& 
			\textbf{96.9$\pm$0.1}&\textbf{96.8$\pm$1.8 }&\textbf{97.1$\pm$0.2} &
			\textbf{80.2$\pm$1.1} & \textbf{76.0$\pm$0.5}& \textbf{62.8$\pm$0.1}\\ 
			\bottomrule
		\end{tabular}
	}	
\end{table}

Table \ref{tab:result} shows a quantitative comparison of fine-tuning performance for the three evaluated downstream tasks. 
From the results of Task I, we observe that AWCL pre-trained models, i.e. \textit{AWCL (coarse)} and \textit{AWCL (fine-grained)}, generally outperform the compared contrastive learning methods  SimCLR and CLPI. 
In particular, \textit{AWCL (coarse)} improves on SimCLR and CLPI by 1.9\% and 3.8\% in F1-score, respectively. 
Compared to the supervised pre-training methods, both AWCL approaches achieve better performance in Recall and F1-score than vanilla supervised pre-training with coarse-grained labels.
These findings suggest that incorporating anatomy information to select positive pairs from multiple scans can notably improve representation learning. 

However, we find that all the contrastive pre-training approaches presented here underperform the supervised pre-training (fine-grained) which has the same form of semantic supervision as Task I. 
This suggests that without explicitly encoding semantic information, contrastively learned representations may provide limited benefits to the generalization of a fine-grained multi-class classification task, which is line with the findings in \cite{Multi-task-CL}.

\subsection{Evaluation on recognition of first-trimester anatomies}
\label{Evaluation-TaskII}

\subsubsection{Evaluation details.} 
We investigate how the pre-trained representations generalize to a cross-domain classification task using the first-trimester US scans. 
This first-trimester classification task recognises five anatomical categories: crown rump length(CRL), nuchal translucency (NT), biparietal diameter (BPD), 3D and background (Bk).
We split the data into training and testing sets (78\%/22\%). 
The trained encoders followed by two fully-connected layers and a softmax layer were fine-tuned for 200 epochs with a learning rate of 0.1 decayed by 0.1 at 150 epochs. The network was trained using SGD with momentum of 0.9. 
Standard data augmentation was used, including rotation $[-30^{\circ}, 30^{\circ}]$, horizontal flip, Gaussian noise, and shear $\leq0.2$. 
Batch size was adjusted according to model size and GPU memory restrictions.
We use the same metrics as presented in Task I for performance evaluation. 
\begin{figure}[t]
	\centering
	\includegraphics[width=12.2cm,height=8cm,keepaspectratio]{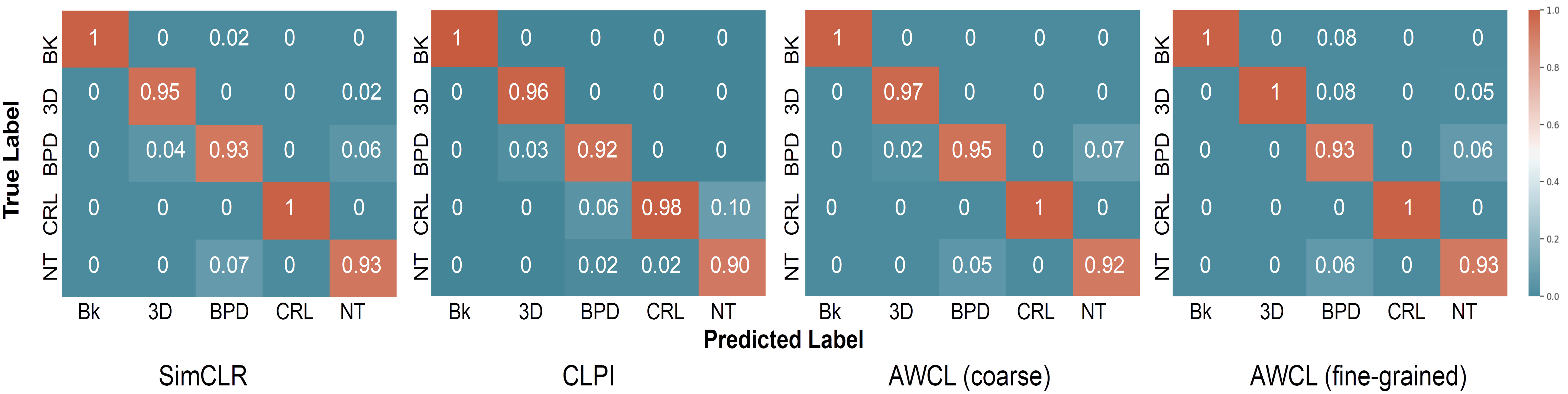}
	\caption{
		Illustration of the confusion matrix for the first-trimester classification task.}
	\label{fig:CM}
\end{figure}

\subsubsection{Results and discussion.} 
For Task II, we see from Table \ref{tab:result}, that \textit{AWCL (fine-grained)} achieves the best performance among all the compared solutions.
In particular, it achieves a performance gain of 4.9\%, 3.4\% and 5.0\% in Precision, Recall and F1-score  compared to ImageNet pre-training, and even improves on supervised pre-training with fine-grained labels (the upper-bound baseline) by 0.7\% in F1-score.
Moreover, \textit{AWCL (coarse)} also surpasses ImageNet and supervised pre-training with coarse-grained labels by 1.9\% and 6.3\% in F1-score.
For comparison with other contrastive learning methods, we observe a similar improved trend as described in Task I, i.e. \textit{AWCL (coarse)} and \textit{AWCL (fine-grained)} perform better than SimCLR and CLPI. Further evidence is provided in Fig. \ref{fig:CM}, which shows that both \textit{AWCL (coarse)} and  \textit{AWCL (fine-grained)} provide better prediction accuracy than CLPI for all anatomy categories.
These experimental results again demonstrate the effectiveness of AWCL approaches and suggest that the inclusion of anatomy information in contrastive learning is a good practice when it is available at hand. 

\subsection{Evaluation on segmentation of NT and CRL}
\subsubsection{Evaluations details.} 
In this section, we evaluate how the pre-trained models generalize to a cross-domain segmentation task with the data from the first-trimester US scans. 
Segmentation of NT and CRL was defined as a three-class segmentation task with the three classes being; mid-sagittal view, nuchal translucency, background. The data is divided into training and testing with 80\%/20\%. We follow the design of ResNet-18 auto-encoder by attaching additional decoders with the trained encoders, and then fine-tuned the entire model for 50k iterations with a learning rate of 0.001, RMSprop optimization (momentum=0.9) and a weight decay of 0.001. 
We apply random scaling, random shifting, random rotation, and random horizontal flipping for data augmentation.
We use global average accuracy (GAA), mean accuracy (MA), and mean intersection over union (mIoU) metrics for evaluating the segmentation task (Task III). 
\begin{figure}[t]
	\centering
	\includegraphics[width=1.\textwidth]{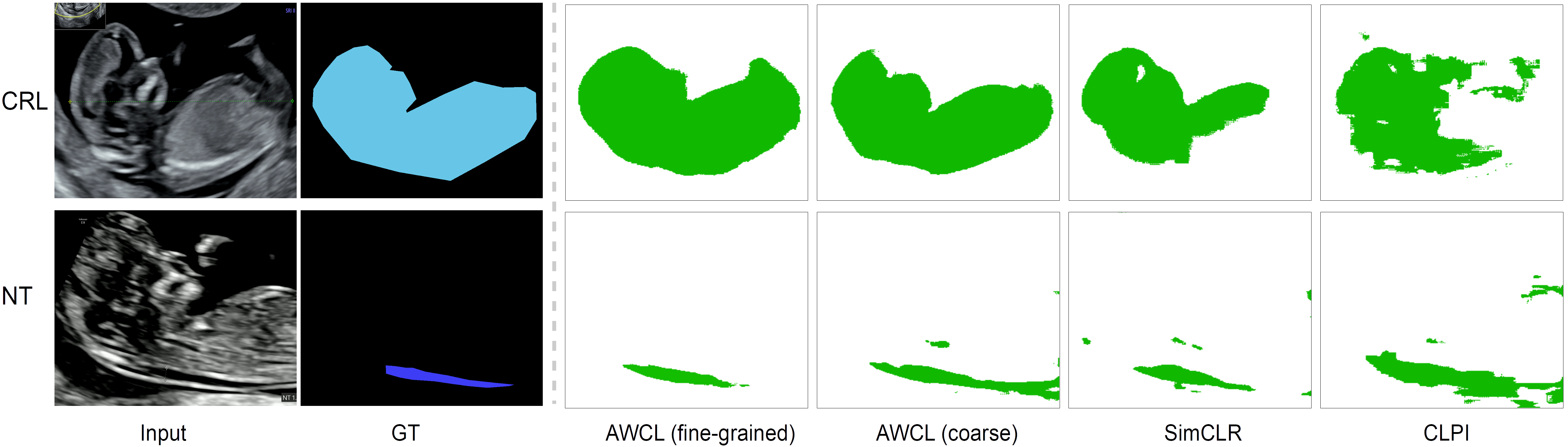}
	\caption{
		Illustration of the qualitative results for the first-trimester segmentation task.
	}
	\label{fig:seg_method}
\end{figure}

\subsubsection{Results and discussion.} 
For Task III, we find that \textit{AWCL (fine-grained)}, achieves comparable or slightly better performance than supervised pre-training with fine-grained labels and surpasses other compared pre-training methods by large margins in mIoU (see Table \ref{tab:result}).
In particular, it outperforms ImageNet and SimCLR by 13.8\% and 7.1\% in mIoU, respectively.
Likewise, \textit{AWCL (coarse)} performs better than ImageNet, supervised pre-training with coarse-grained labels, SimCLR and CLPI by large margins in most evaluation metrics.
Fig. \ref{fig:seg_method} also visualizes the superior performance of  \textit{AWCL (fine-grained)} and \textit{AWCL (coarse)} compared to SimCLR and CLPI, which aligns with the quantitative evaluation.
These observations suggest that AWCL are able to learn more meaningful semantic representations that are beneficial for this pixel-wise segmentation task. 
Overall, the evaluated results on Tasks II and III  demonstrate that the AWCL models report consistently better performance than the compared pre-trained models, implying the advantage of learning task-agnostic features that can better generalized to the tasks from different domains.

\section{Analysis}
\label{Ablation study}
\subsection{Partial fine-tuning} 
\label{Linear Evaluation}

To analyze representation quality, 
we extract fixed feature representations from the last layer of the ResNet-18 encoder and then evaluate them in two classification target tasks (Task I and Task II).
Experimentally, we freeze the entire backbone encoder and attach a classification head \cite{sononet} for Task I, and a non-linear classifier as mentioned in Section \ref{Evaluation-TaskII} for Task II. 
From Table \ref{tab:result-linear}, we observe that the AWCL approaches show better representation quality by surpassing the three compared approaches in terms of F1-score for both tasks. 
This suggests that the learned representations are strong non-linear features which are more generalizable and transferable to the downstream tasks.
Comparing Tables  \ref{tab:result} and  \ref{tab:result-linear}, we find that although the reported scores of partial fine-tuning are generally lower than for full fine-tuning, the performance between two implementations of transfer learning  is correlated.

\begin{table}
	\caption{\footnotesize Performance comparison of partial fine-tuning  (mean±std.[\%]) on the tasks of standard plane detection (Task I) and first-trimester anatomy recognition (Task II). Best results are
		marked in \textbf{bold.}}
	\centering
	\label{tab:result-linear}
	\resizebox{0.85\linewidth}{!}{
		\begin{tabular}{l|ccc|ccc}
			\toprule
			&\multicolumn{3}{@{}c@{}}{Task I}&   \multicolumn{3}{@{}c@{}}{Task II}\\
			\hline
			Pre-training methods  &Precision ($\uparrow$) & Recall ($\uparrow$) &   F1-score ($\uparrow$) &   Precision ($\uparrow$)&   Recall ($\uparrow$)&   F1-score ($\uparrow$)
			\\
			\hline

			ImageNet &65.5$\pm$4.9 &58.1$\pm$4.3  &60.2$\pm$4.9&	84.03$\pm$0.13&84.25$\pm$0.45 &83.92$\pm$0.62	\\

			SimCLR &67.6$\pm$3.5 & 67.3$\pm$4.1  &65.9$\pm$3.6&  87.65$\pm$0.09&86.82$\pm$0.11 &86.12$\pm$0.20\\
			
			CLPI &\textbf{71.2$\pm$0.6} &68.6$\pm$4.9 &67.9$\pm$5.1 &82.07$\pm$0.62&80.28$\pm$0.83 &81.10$\pm$1.03  \\	
			
			\hline
			\textit{AWCL (coarse)} &70.5$\pm$2.7& \textbf{71.3$\pm$1.7 }&\textbf{69.5$\pm$1.9}&86.21$\pm$1.20&87.67$\pm$0.32 &86.14$\pm$0.59   \\
			
			\textit{AWCL (fine-grained)} &69.7$\pm$1.0 &68.8$\pm$0.2   & 68.4$\pm$0.5& \textbf{88.65$\pm$0.49}&\textbf{88.14$\pm$0.17} &\textbf{87.00$\pm$0.01}\\ 
			\bottomrule
		\end{tabular}
	        }	
\end{table}

\subsection{Visualization of feature representations} 
\label{Visualization analysis}
\begin{figure}[t]
	\centerline{\includegraphics[width=\columnwidth]{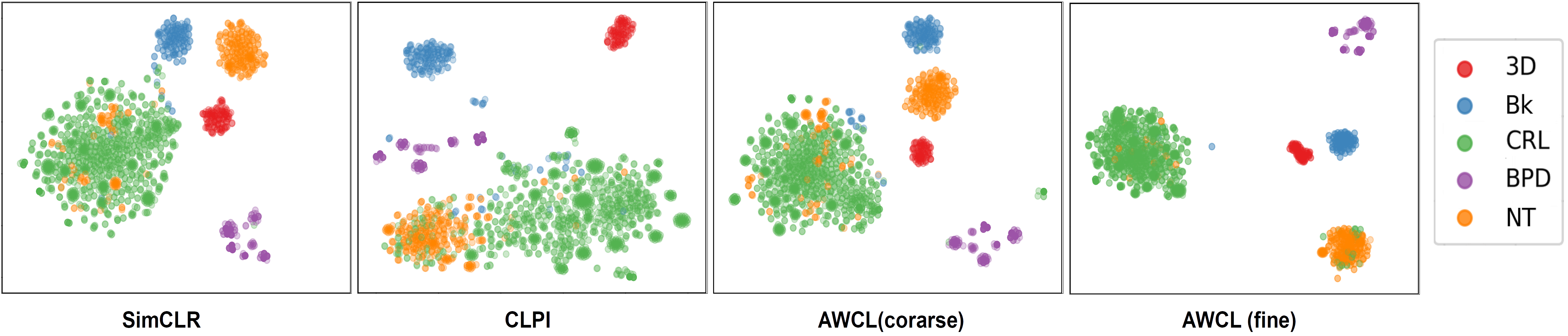}}
	\caption{t-SNE feature visualization of the model penultimate layers on Task II.}
	\label{fig-TSNE}
\end{figure}

In this section we investigate why the feature representations produced with AWCL pre-trained models result in better downstream task performance.
We visualize the image representations of Task II extracted from the penultimate layers using t-SNE \cite{tSNE}  in Fig. \ref{fig-TSNE}, where different anatomical categories are denoted with different color. 
We compare the resulting t-SNE embeddings of AWCL models with those as SimCLR and CLPI.
We observe that the feature representation by CLPI is not quite separable, especially for classes of NT (\textcolor{orange}{orange}) and CRL (\textcolor{green}{green}).
The features embeddings from SimCLR are generally better separated than those in CLPI, while confusion between CRL (\textcolor{green}{green}) and Bk (\textcolor{cyan}{blue}) remains. 
By comparison, \textit{AWCL (fine-grained)} achieves the best separated clusters among five anatomical categories, which means that the learned representations in the embedding space are more distinguishable.
These visualization results demonstrate that AWCL approaches are able to learn discriminative feature representations which are better generalized to downstream tasks.

\subsection{Impact of data granularity on AWCL} 
\label{Analysis of Data granularity in AWCL}
\begin{figure}[t]
\centerline{\includegraphics[width=0.65\columnwidth]{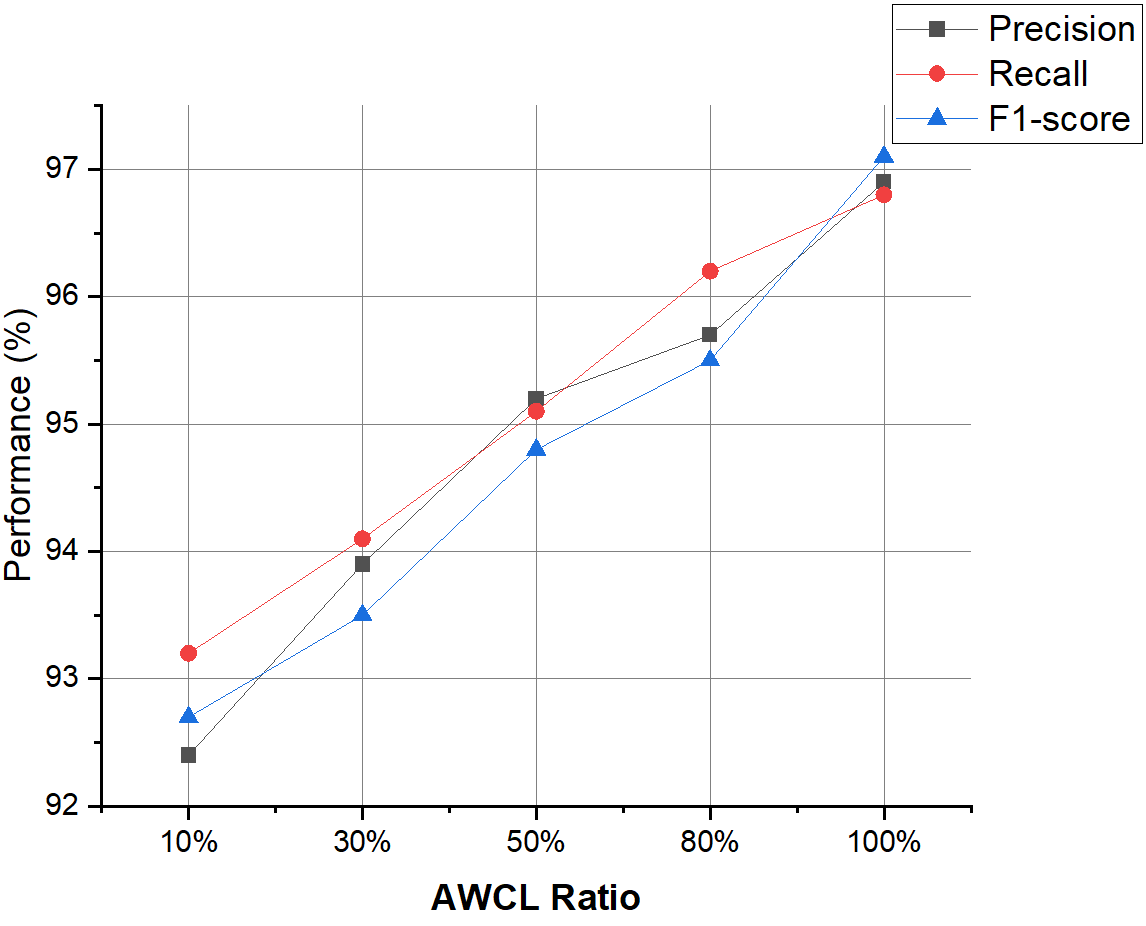}}
\caption{Impact of anatomy ratio on AWCL (fine-grained) evaluated on Task II.}
\label{fig-anatomy-ratio}
\end{figure}

We analyze how the inclusion of coarse- and fine-grained anatomy information impact the AWCL framework, by comparing the experimental results  between \textit{AWCL (coarse)} and \textit{AWCL (fine-grained)} from Section \ref{Evaluation on Standard Plane Detection} to Section \ref{Visualization analysis}.
Based on the transfer learning results in Table \ref{tab:result}, 
we find that \textit{AWCL (fine-grained) } achieves better performance than \textit{AWCL (coarse)} for Tasks II and III, despite the slight performance drop in Task I. We hypothesize that \textit{AWCL (coarse)} learns more generic representations than \textit{AWCL (fine-grained)}, which leads to better in-domain generalization.
Qualitative results in Fig. \ref{fig:CM} and Fig. \ref{fig:seg_method} also reveal the advantage of \textit{AWCL (fine-grained)} over its counterpart.
Based on the ablation analysis, Table \ref{tab:result-linear} shows a similar finding as seen in Table \ref{tab:result}.
Fig. \ref{fig-anatomy-ratio}  shows that feature embeddings of \textit{AWCL (fine-grained)} are more discriminative than those of \textit{AWCL (coarse)} thereby resulting in better generalization to downstream tasks.
These observations suggest the importance of learning intra-class feature representations for better generalization to downstream tasks especially when there is a domain shift.

\subsection{Impact of anatomy ratio on AWCL} 
\label{Impact of label fraction in AWCL}
We investigate how varying anatomy ratios impact the AWCL framework. 
Note that higher anatomy ratio represents that larger number of samples from same or different US scans belonging to the same anatomy category are included to form positive pairs for contrastive learning.
We incorporated the anatomy information with four different ratios: 10\%, 30\%, 50\%, and 80\% to train the AWCL (fine-grained) models on the pre-training dataset, respectively. Then, we evaluate these trained models on Task II via full fine-tuning.
As shown in Fig. \ref{fig-anatomy-ratio}, we observe that the performance improves with an increasing anatomy ratio. 
It suggests that using more distinct but anatomically similar samples to compose positive pairs results in a better quality representation.

\section{Conclusion}
\label{Conclusion}
In this paper, we presented a new anatomy-aware contrastive learning (AWCL) approach for fetal ultrasound imaging tasks. The proposed approach is able to leverage more positive samples from the same or different US videos with the same anatomy category and align well with the anatomical characteristics of ultrasound videos.
The feature representative analysis shows AWCL approaches learn discriminative representations that can be better generalized to downstream tasks.
Through the reported comparative study, AWCL with fine-grained anatomy information which preserves intra-class difference was more effective than its counterpart. 
Experimental evaluations demonstrate that our AWCL approach provides useful transferable representations for various downstream clinical tasks, especially for cross-domain generalization.
The proposed approach can be potentially applied to other medical imaging modalities where such anatomy information is available. 

\subsection*{Acknowledgement}
The authors would like to thank Lok Hin Lee, Richard Droste, Yuan Gao and Harshita Sharma for their help with data preparation.
This work is supported by the EPSRC Programme Grants Visual AI (EP/T028572/1) and Seebibyte (EP/M013774/1), the ERC Project PULSE (ERC-ADG-2015 694581), the NIH grant U01AA014809, and the NIHR Oxford Biomedical Research Centre. The NVIDIA Corporation is thanked for a GPU donation.

\bibliographystyle{splncs04}
\bibliography{AWCL}

\end{document}